\documentclass{Interspeech}

\interspeechcameraready

\title{How to Retrieve Examples in In-context Learning to Improve Conversational Emotion Recognition using Large Language Models?}

\author[affiliation={1}]{Mengqi}{Wang}
\author[affiliation={1}]{Tiantian}{Feng}
\author[affiliation={1}]{Shrikanth}{Narayanan}

\affiliation{}{University of Southern California}{USA}
\email{}

\usepackage{comment}
\usepackage{xcolor}
\usepackage{booktabs}
\usepackage[export]{adjustbox}
\definecolor{darkgreen}{rgb}{0.0, 0.6, 0.0}

\begin{document}

\maketitle

\begin{abstract}
    Large language models (LLMs) have enabled a wide variety of real-world applications in various domains. However, creating a high-performing application with high accuracy remains challenging, particularly for subjective tasks like emotion recognition. Inspired by the SLT 2024 GenSER Challenge, this study investigates approaches to improving conversational emotion recognition (CER) by LLMs. Specifically, we explore how to retrieve high-quality examples in in-context learning  (ICL) to enhance CER. We propose various strategies based on random and augmented example retrieval and also analyze the impact of conversational context on CER accuracy. Experiments were conducted on the three datasets including IEMOCAP, MELD and EmoryNLP. The results show that augmented example retrieval consistently outperforms other techniques under investigation across all datasets, highlighting the importance of retrieving coherent targeted examples and enhancing them through paraphrasing. 
\end{abstract}

\section{Introduction}

\begin{figure}[htb]
    \begin{minipage}[b]{1\linewidth}
        \centering
        \centerline{\includegraphics[width=8.5cm]{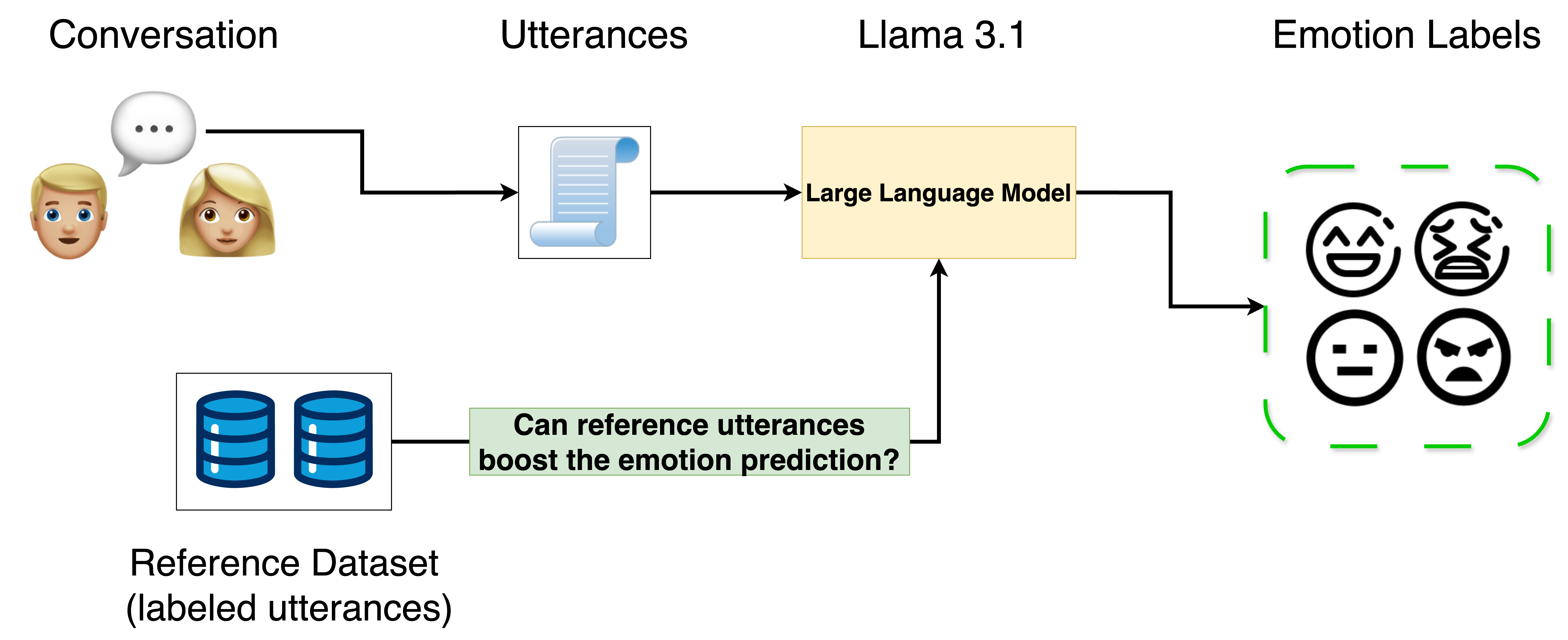}}
            \caption{Overview of our framework: Our goal is to use the LLMs to classify emotions from the utterances in the conversation. Our investigation focuses on evaluating whether the incorporation of reference utterances alongside the target utterances improves the emotion prediction by LLMs.}
        \label{fig:1.1}
    \end{minipage}
    \vspace{-3mm}
\end{figure}

The advancement in artificial intelligence based on deep learning has catalyzed the development of progressively sophisticated large language models (LLMs) capable of understanding and interpreting human context \cite{lecun2015deep}. This in turn has facilitated a wide range of applications, including conversational agents, document processing, and educational tools. These varied applications of LLMs have garnered considerable interest in exploring their abilities to acquire ``human-like" skills. While LLMs have demonstrated skills akin to those of humans, their ability to generate accurate and consistent results requires further investigation to ensure their robustness across various tasks, especially in subjective reasoning tasks such as emotion recognition in conversations. Conversational emotion recognition (CER) involves extracting and analyzing contextual and emotional cues from speech or language contexts to infer human emotions. Recent studies \cite{inproceedings, li2024revise, kyung2024enhancing, wu2024beyond, feng2024foundation} have investigated various strategies to enhance the performance of LLMs in emotion prediction.  In this paper, inspired by the LLM-Based Post ASR Speech Emotion Recognition Challenge from SLT 2024 GenSER Challenge \cite{10832176}, we investigate how to retrieve examples in in-context learning (ICL) to improve CER using the LLMs, as shown in Figure~\ref{fig:1.1}. Our study is based on three distinct datasets: IEMOCAP \cite{busso2008iemocap}, MELD \cite{poria2019meldmultimodalmultipartydataset}, and EmoryNLP Emotion Prediction Dataset \cite{zahiri2018emotion}. 

One closely related work is by Santoso et al. \cite{10448316}, which incorporates conversational context and acoustic features into the prompt to improve the CER performance of GPT-3.5-turbo\footnote{https://platform.openai.com/docs/models/gpt-3-5}. They conducted two experiments: one without conversational context (zero-shot) and the other with a maximum of 20 history utterances included (zero-shot with context). Then, the acoustic features (speaking rate, pitch, intensity, etc.) were extracted and categorized as low, normal, and high based on the thresholds derived from the lowest 30\% and highest 30\% of the numerical values of those features. The evaluations indicated that including conversation segments and acoustic features substantially improves performance compared to single utterances.

Despite this, there are some limitations tied to this work. First, the data confidentiality cannot be guaranteed. The GPT-3.5-turbo is utilized as a cloud-based model, and their prompt includes real speakers' utterances, potentially posing a risk to data privacy. Instead, we mitigate this uncertainty in our study by employing locally deployable open-source models, like the Llama-3 model family \cite{dubey2024llama3herdmodels}. Second, the study exclusively relied on one dataset, which restricts the generalization of the results. We address this issue by examining the patterns in three datasets.

\begin{figure*}[ht]
    \centering
    \includegraphics[width=0.99\textwidth,keepaspectratio]{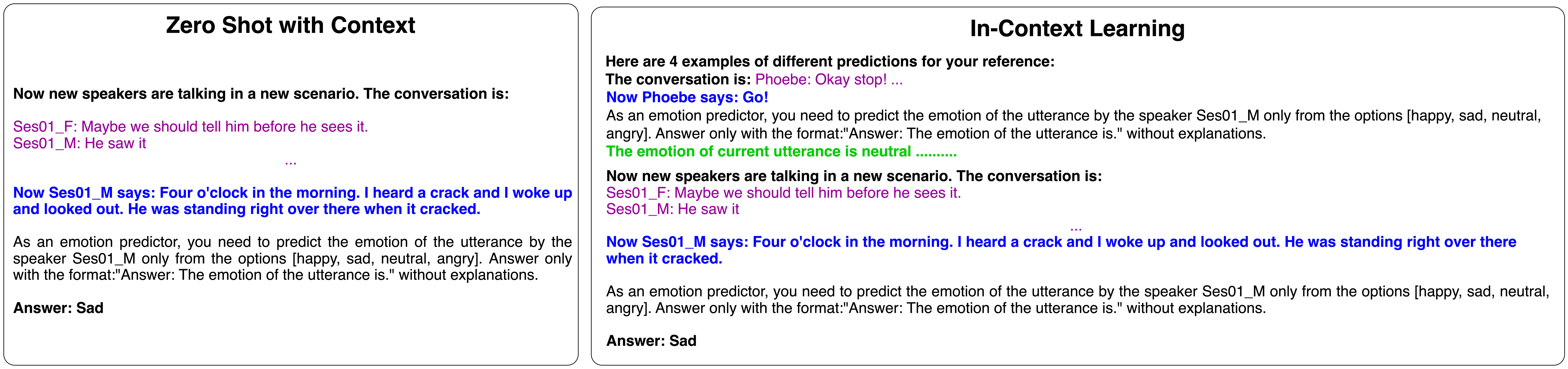}
    \caption{Prompts for zero-shot with context and in-context learning. The purple color is the conversation context, the blue color is the target utterance, and the green color is the answers for reference utterances.}
    \label{fig:2.1}
    \vspace{-3mm}
\end{figure*}

Another related study was conducted by Wu et al. \cite{wu2022self} about ICL \cite{min2022rethinking}. They introduced self-adaptive in-context learning, which is a general select-then-rank framework supported by Top-K selection \cite{liu2021makes} and the Minimal Description Length (MDL) principle. The Top-K selection filters out a smaller candidate set based on the highest semantic similarity and then decides on the best examples by ranking them with the MDL principle. Compared with prompting without in-context examples, ICL methods generally increase the performance of the language modeling, while the performance is inconsistent with random examples. The proposed self-adaptive ICL method can significantly outperform standard approaches in ICL. We note that our study on conversational emotion recognition using ICL is inspired by the TopK approach proposed in \cite{wu2022self}. In summary, the main contributions and findings of this paper are summarized below:

\begin{itemize}
    \item First, we investigate how the size of the conversation context impacts the emotion recognition performance and find that expanding the context does not always increase the performance. Specifically, the performance starts to saturate beyond a certain context size.
    
    \item Second, we explore creating a reference dataset by combining MELD and EmoryNLP used for the baseline ICL. The baseline ICL is performed by providing a set of randomly selected annotated emotional utterances in each prediction. However, giving random utterances as ICL examples does not yield consistent performance improvements.
    
    \item We develop an ICL framework, the Augmented Example Retrieval (AER), which aims to improve the performance of LLMs by selecting the most coherent example from a reference dataset based on similarity measures computed using SentenceTransformer \cite{reimers2019sentencebertsentenceembeddingsusing}. Our results show that AER consistently improves macro F1 across all three test datasets.
    
    \item We finally find that AER can be adapted to noisy text data obtained from Automatic Speech Recognition (ASR) transcriptions, yielding consistent performance improvements compared to zero-shot baselines.
\end{itemize}

\section{Methods}

Our experiment is designed as three steps. First, we evaluate the zero-shot baseline by providing examples in the prompt. Further, we study how the size of the conversation context impacts zero-shot emotion recognition. Next, we perform the ICL baseline experiment by including external reference datasets. Specifically, we construct the prompt by incorporating randomly selected example utterances. Finally, we refine the prompt by choosing more coherent examples for the LLMs with more deterministic cues. Figure~\ref{fig:2.1} shows the prompt templates for different methods.

\begin{figure*}[ht]
    \centering
    \centerline{\includegraphics[width=\textwidth]{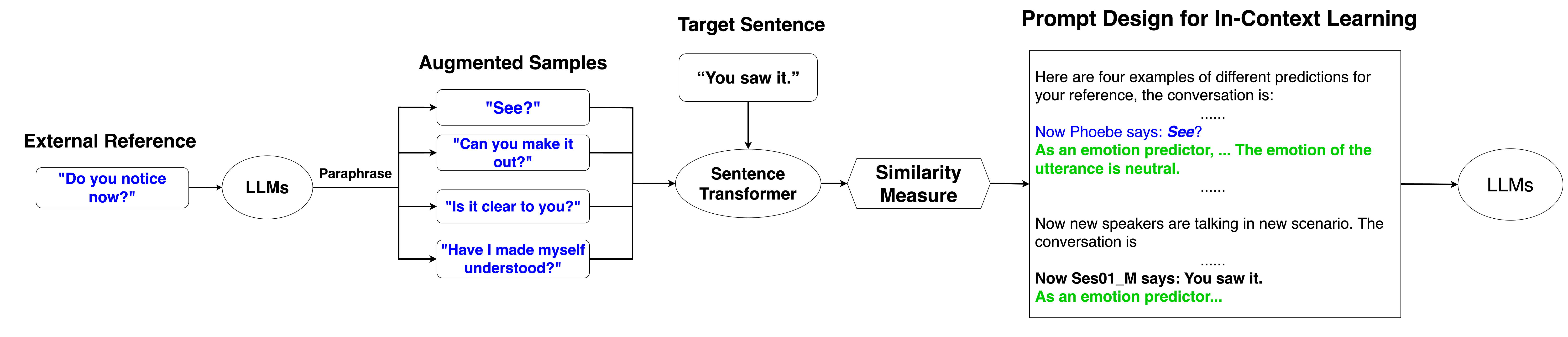}}
    \caption{Flow of our proposed ICL with Augmented Example Retrieval (AER). Given reference utterances, we prompt LLM to generate four paraphrases for each one and encode them with SentenceTransformer. Every time we have a target utterance during prediction, we encode it and find the reference utterance with the highest cosine similarity, while selecting another three examples with the remaining emotions arbitrarily. We design prompts for in-context learning accordingly, as shown above.}
    \label{fig:2.4}
    \vspace{-3mm}
\end{figure*}

\subsection{Zero-Shot Baseline}
Given that the conversation context may provide relevant information for LLMs to predict the emotions, our prompt followed the SLT GenSER challenge to include previous utterances of the target utterance. The conversation context size is defined as the number of preceding utterances included in the prompt prior to the target utterance. Specifically, we experiment with the conversation context in $\{0, 5, 10, 15, 20\}$.  Note that a conversation context size of $0$ amounts to the standard zero-shot learning.

\begin{table}
    \centering
    \begin{tabular}{ccccc}
        \toprule
        Dataset & happy & sad & neutral & angry \\
        \midrule
        IEMOCAP & 681 & 475 & 607 & 814\\
        MELD & 402 & 208 & 1256 & 345\\
        EmoryNLP & 282 & 98 & 349 & 113\\
        \bottomrule
    \end{tabular}
    \vspace{2mm}
    \caption{The number of annotated emotional utterances across all three test datasets}
    \label{stats}
\end{table}


\subsection{ICL with Random Examples}
Following the zero-shot experiments, either with or without conversation contexts, we further investigate methods to improve the emotion prediction made by the LLMs. Specifically, we focus on in-context learning where prompts are constructed using examples from the combined reference dataset from MELD and EmoryNLP. Here, the reference dataset includes only the training subsets of these two datasets. Unlike conventional ICL, where the examples are directly incorporated into the prompt, we augment the dataset by generating distinct paraphrases. Subsequently, we form the prompt by selecting the examples from the augmented reference dataset. We employ the Mistral-7B model \cite{jiang2023mistral7b} for rephrasing. When designing the final prompt, we randomly selected four example utterances, each associated with a distinct emotion from the public dataset.

\subsection{ICL with Augmented Example Retrieval (AER)}

We introduced a more advanced technique to construct the prompt to the LLM through coherent example retrieval. Specifically, we build the knowledge database by encoding utterances from public datasets using the SentenceTransformer. We empirically investigate the retrieval from the in-domain and out-of-domain data sources. By retrieving the more coherent examples of the targeted utterance, we aim to design a better prompt incorporating more relevant context for the prediction. Overall, our technique of Augmented Example Retrieval (AER), as illustrated in Figure~\ref{fig:2.4}, is described below: 

We employ SentenceTransformer \cite{reimers2019sentencebertsentenceembeddingsusing} to encode all the utterances in the reference dataset. For each target utterance that requires emotion prediction, after encoding, we search for the example utterance based on the cosine similarity. We record its four paraphrases, while the other three examples with the remaining emotions are chosen arbitrarily. We carry out five prediction rounds on the same public dataset and prompt a distinct paraphrase of the most coherent example in each round while keeping the other examples unchanged. The results are determined by the majority votes after five rounds.

\section{Datasets and Models}
\subsection{SLT Baseline}
This work starts with the SLT 2024 GenSEC Challenge. The objective of the challenge is to scrutinize and enhance the accuracy of emotion perception by large language models (LLMs) with ICL. Currently, it focuses on the prediction of four emotions \textbf{[happy, sad, neutral, angry]}. We evaluate the performance of the prediction with macro F1.

\subsection{Test Datasets for the ICL}
We utilized the Interactive Emotional Dyadic Motion Capture (IEMOCAP) \cite{busso2008iemocap} dataset provided by the Challenge, which contains the scripts from distinct speakers, with both ground-truth version and automatic speech recognition (ASR) versions provided. In addition, we consider two public datasets, Multimodal EmotionLines Dataset (MELD) \cite{poria2019meldmultimodalmultipartydataset} test data and EmoryNLP Emotion Prediction \cite{zahiri2018emotion} test data. The criterion of our selection is that the utterance will be considered to predict if its emotion is in [happy, sad, neutral, angry]. The statistics of the datasets are given in Table~\ref{stats}.

\noindent \textbf{IEMOCAP} \cite{busso2008iemocap} contains five sessions and each one has a conversation between a male speaker and a female speaker. It has 2577 valid utterances to be predicted. There are groundtruth version and eleven ASR versions for each utterance and we specifically inspect '\textit{Groundtruth}', '\textit{Hubertlarge}' \cite{hsu2021hubert}, '\textit{W2v2100}' \cite{baevski2020wav2vec}, and '\textit{Whispertiny}' \cite{radford2023robust} transcripts in our experiments.

\begin{figure}
    \begin{minipage}[b]{1\linewidth}
        \centering
        \centerline{\includegraphics[width=8.5cm]{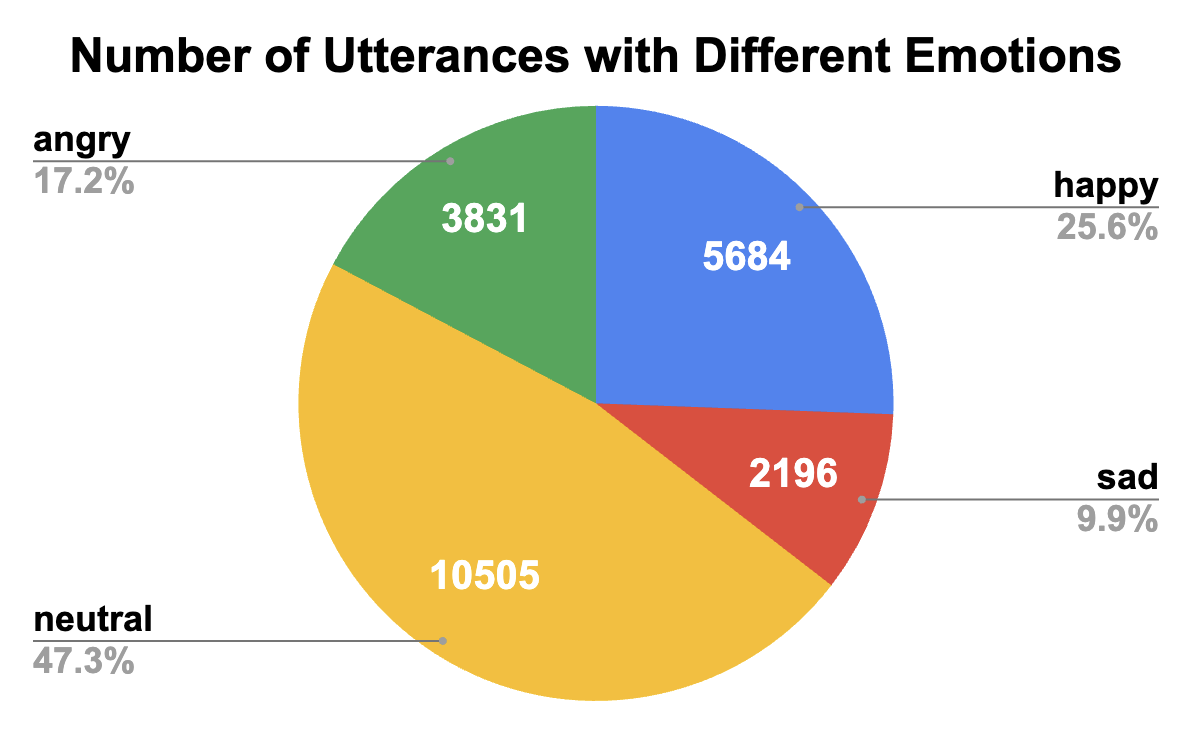}}
        \caption{The statistics of the augmented reference dataset.}
        \label{reference_dataset}
    \end{minipage}
    \vspace{-3mm}
\end{figure}

\noindent \textbf{MELD} \cite{poria2019meldmultimodalmultipartydataset} is referenced from Declare-Lab and consists of utterances from multiple speakers from Friends TV series. We only focus on predicting the utterances with emotion in [Joy, Sadness, Neutral, Anger], which are converted to [happy, sad, neutral, angry]. After the data processing, we predict the emotions of 2211 utterances within the group. For the reference dataset, we utilize 8245 utterances from the training data.

\noindent \textbf{EmoryNLP} \cite{zahiri2018emotion} is adapted from the public emotion detection data from EmoryNLP group from Emory University. There are 7 emotions in this dataset and we only focus on utterances with emotion in [joyful, sad, neutral, mad] which is mapped to [happy, sad, neutral, angry]. After data processing, the number of utterances we experiment with inlcudes 842 utterances from the test data and 6965 utterances from training data for the reference dataset.

The rest of the utterances that do not comply with the standard will be incorporated into the conversation context. We also apply data augmentation on the reference dataset to form an augmented reference dataset. The statistics of the dataset are shown in Figure~\ref{reference_dataset}.




\subsection{Models}
\textbf{Large Language Models (LLMs)}. We use Mistral-7B-Instruct-v0.2 \cite{jiang2023mistral7b} to carry out data augmentation. In addition, we use Llama-3.1-8B-Instruct \cite{dubey2024llama3herdmodels} in our experiments. Both of these were downloaded from huggingface\footnote{https://huggingface.co/}. The data type for tensor data for these LLMs is Brain Floating Point 16 (bfloat16). The models are run with a temperature 0.0001. The dimensions of these LLMs are shown in Table~\ref{dim}.

\noindent \textbf{Text Embedding Model}. We use Sentence-Transformer-all-MiniLM-L6-v2 \cite{reimers2019sentencebertsentenceembeddingsusing}, imported from the SentenceTransformer package, to encode utterances. It maps the sentences to a dense vector space of dimension 384.

\begin{table}
    \centering
    \begin{tabular}{cc}
    \toprule
    Model & Dimension \\
    \midrule
    Mistral-7B-Instruct-v0.2 & 4096 \\
    Meta-Llama-3.1-8B-Instruct & 4096 \\
    \bottomrule
    \end{tabular}
    \vspace{2mm}
    \caption{The dimension of the LLMs}
    \label{dim}
\end{table}

\section{Results}

\subsection{Baselines - Zero-shot without Examples}
\label{ssec:subhead}

\begin{figure}
    \begin{minipage}[b]{1\linewidth}
        \centering
        \centerline{\includegraphics[width=8.5cm]{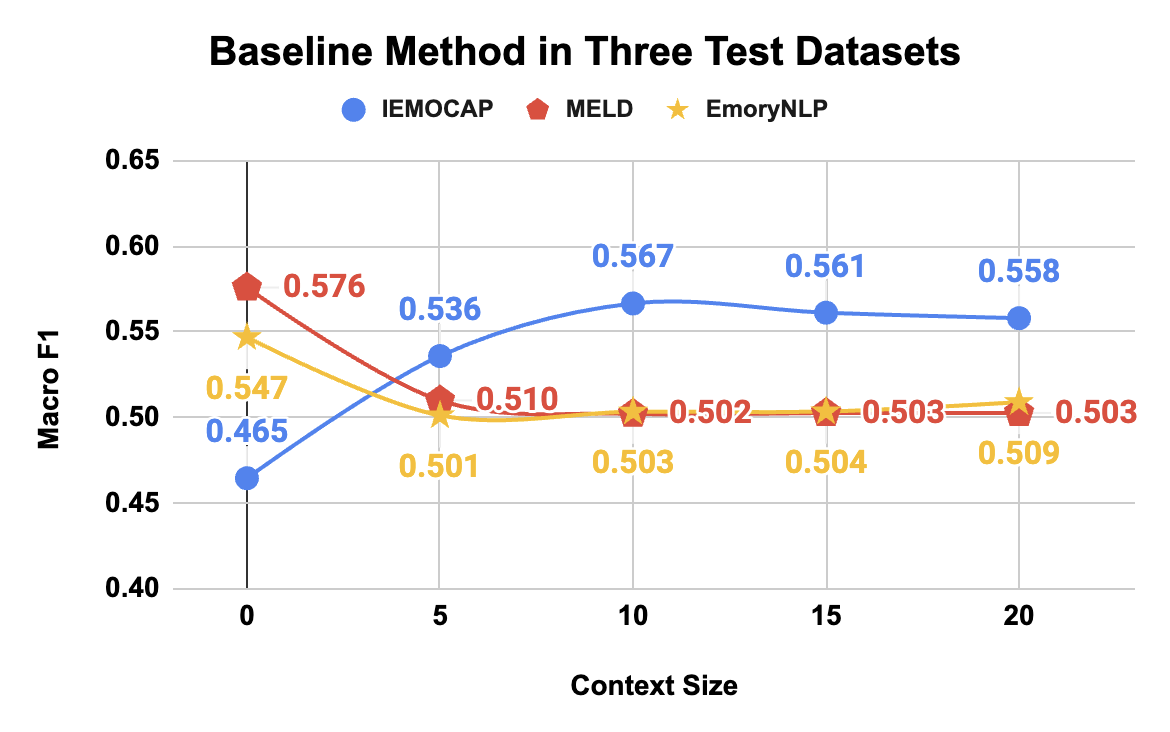}}
        \caption{The fluctuation of macro F1 in different conversation context size for IEMOCAP, MELD, and EmoryNLP datasets.}
        \label{4.1}
    \end{minipage}
    \vspace{-3mm}
\end{figure}

The results reveal different prediction behaviors between IEMOCAP and the other two datasets from Figure~\ref{4.1}. On IEMOCAP, we observe that an increase in the conversation context size tends to improve the macro F1 score. Specifically, there is a significant improvement of performance from 0.465 to 0.536 when the conversation context size expands from 0 to 5. This increase in the performance saturated to around 0.567 once the conversation context size reaches 10. However, the LLM achieves the highest performance with no conversation context provided on the other two datasets, 0.576 in MELD and 0.547 in EmoryNLP. The macro F1 decreases to approximately 0.5 as the size of conversation context increases to 10. Based on the results, we  continue our experiments on the conversation context size with best results across three datasets. 

\subsection{ICL with Random Examples}
\label{ssec:subhead}
According to the results in Table~\ref{prompt_technique}, the incorporation of randomly selected examples from the augmented reference dataset does not consistently boost the overall performance in all three datasets. For IEMOCAP and MELD, the macro F1 is comparable with the baseline results, while there is about 2\% improvement on the MELD dataset. 

\subsection{ICL with AER Results}
As shown in Table~\ref{prompt_technique}, we observe a clear improvement of macro F1 in all three test datasets with the integration of AER, compared with the baseline and the ICL with random examples. On average, we see around $1\%$ improvement across all three datasets, and particularly over $2\%$ in the EmoryNLP dataset, compared with the baselines.
\begin{table}[h]
    \centering
    \small
    \begin{tabular}{cccc}
        \toprule
         & IEMOCAP & MELD & EmoryNLP \\
        \midrule
        Baseline & \textbf{0.567} & \textbf{0.576} & \textbf{0.547} \\
        ICL-Random Examples & \textbf{0.568} & \textbf{0.575} & \textbf{0.566} \\
        ICL-AER &  \textbf{\textcolor{darkgreen}{0.575}} & \textbf{\textcolor{darkgreen}{0.581}} & \textbf{\textcolor{darkgreen}{0.570}}\\
        \bottomrule
    \end{tabular}
    \vspace{2mm}
    \caption{Macro F1 of the baseline, ICL-Random Examples, and the ICL-AER. The context sizes for IEMOCAP, MELD, and EmoryNLP are 10, 0, and 0.}
    \label{prompt_technique}
    \vspace{-3mm}
\end{table}

\subsection{ICL with AER on ASR Transcription}
Given the improvement of CER with AER, we also achieve more accurate results after using noisy ASR transcripts, relative to the baseline method. As seen in Table~\ref{asr_table}, the accuracy improves in all the ASR transcripts cases.

\begin{table}[h]
    \centering
    \footnotesize
    \begin{tabular}{ccccc}
        \toprule
         & Groundtruth & Hubertlarge & W2v2100 & Whispertiny \\
        \midrule
        Baseline & 0.567 & 0.532 & 0.503 & 0.487 \\
        AER & \textbf{\textcolor{darkgreen}{0.575}} & \textbf{\textcolor{darkgreen}{0.558}} & \textbf{\textcolor{darkgreen}{0.532}} & \textbf{\textcolor{darkgreen}{0.489}}\\
        \bottomrule
    \end{tabular}
    \vspace{2mm}
    \caption{Macro F1 of different ASR transcripts in Baseline and ICL-AER. The context size is 10.}
    \label{asr_table}
\end{table}

\subsection{Discussion}
In our experimental settings, we found that the larger conversation context size can not consistently improve  CER across all three datasets examined. We categorize other predicted emotions to \textbf{neutral}, and MELD and EmoryNLP datasets have relatively imbalanced emotion distribution, where utterances with neutral emotion are significantly more prevalent than the others. We speculate that the LLM tends to respond with more diverse emotions that are not in our emotion list, which are treated as \textbf{neutral}, as less conversation context provided. In addition, when we retrieve random examples for the LLM, it may introduce bias which induces inaccurate prediction, as reflected by our results.

\section{Conclusion}

In this work, we study how to retrieve examples in ICL to improve emotion recognition in conversations using LLMs. We specifically proposed an augmented example retrieval approach to prompt the LLMs with the most coherent example to the target utterance. Our experiments show that ICL with randomly selected examples performs comparable to baseline zero-shot learning. Moreover, our proposed AER can effectively improve emotion recognition performance over zero-shot learning and ICL with randomly chosen examples. The performance improvement is consistent across all datasets and text input sources (e.g., ASR transcript, human transcript). 

\section{Limitations and Future Work}
Despite the promise of the proposed AER method, there are limitations that need further refinement. $1)$ Due to the restriction of GPU capacity, we chose to only experiment on Llama-3.1-8B-Instruct; more complex models should be investigated. $2)$ The test datasets are limited and cannot assure the generalization of our method. For example, some utterances in the test datasets do not have accurate emotion labels, and some sessions do not have enough conversation context to support the prediction. In future work, we will extend our experiments to more diverse LLMs, such as Gemma-2 Family \cite{team2024gemma}. We will also incorporate other richer public datasets, such as MSP-Improv \cite{busso2016msp}, and perform data screening and processing on the datasets to be more accurate and complete. Additionally, we also aim to refine the method to enhance the conversational emotion recognition to a wider spectrum of expressed emotions.

\bibliographystyle{IEEEtran}
\bibliography{mybib}

\begin{thebibliography}{10}
\providecommand{\url}[1]{#1}
\csname url@samestyle\endcsname
\providecommand{\newblock}{\relax}
\providecommand{\bibinfo}[2]{#2}
\providecommand{\BIBentrySTDinterwordspacing}{\spaceskip=0pt\relax}
\providecommand{\BIBentryALTinterwordstretchfactor}{4}
\providecommand{\BIBentryALTinterwordspacing}{\spaceskip=\fontdimen2\font plus
\BIBentryALTinterwordstretchfactor\fontdimen3\font minus \fontdimen4\font\relax}
\providecommand{\BIBforeignlanguage}[2]{{%
\expandafter\ifx\csname l@#1\endcsname\relax
\typeout{** WARNING: IEEEtran.bst: No hyphenation pattern has been}%
\typeout{** loaded for the language `#1'. Using the pattern for}%
\typeout{** the default language instead.}%
\else
\language=\csname l@#1\endcsname
\fi
#2}}
\providecommand{\BIBdecl}{\relax}
\BIBdecl

\bibitem{lecun2015deep}
Y.~LeCun, Y.~Bengio, and G.~Hinton, ``Deep learning,'' \emph{nature}, vol. 521, no. 7553, pp. 436--444, 2015.

\bibitem{inproceedings}
H.~Wu, H.-C. Chou, K.-W. Chan, L.~Goncalves, J.~Du, J.-S. Jang, C.-C. Lee, and H.-y. Lee, ``Empower typed descriptions by large language models for speech emotion recognition,'' 09 2024.

\bibitem{li2024revise}
Y.~Li, Y.~Gong, C.-H.~H. Yang, P.~Bell, and C.~Lai, ``Revise, reason, and recognize: Llm-based emotion recognition via emotion-specific prompts and asr error correction,'' \emph{arXiv preprint arXiv:2409.15551}, 2024.

\bibitem{kyung2024enhancing}
J.~Kyung, S.~Heo, and J.-H. Chang, ``Enhancing multimodal emotion recognition through asr error compensation and llm fine-tuning,'' in \emph{Proc. Interspeech 2024}, 2024, pp. 4683--4687.

\bibitem{wu2024beyond}
Z.~Wu, Z.~Gong, L.~Ai, P.~Shi, K.~Donbekci, and J.~Hirschberg, ``Beyond silent letters: Amplifying llms in emotion recognition with vocal nuances,'' \emph{arXiv preprint arXiv:2407.21315}, 2024.

\bibitem{feng2024foundation}
T.~Feng and S.~Narayanan, ``Foundation model assisted automatic speech emotion recognition: Transcribing, annotating, and augmenting,'' in \emph{ICASSP 2024-2024 IEEE International Conference on Acoustics, Speech and Signal Processing (ICASSP)}.\hskip 1em plus 0.5em minus 0.4em\relax IEEE, 2024, pp. 12\,116--12\,120.

\bibitem{10832176}
C.-H.~H. Yang, T.~Park, Y.~Gong, Y.~Li, Z.~Chen, Y.-T. Lin, C.~Chen, Y.~Hu, K.~Dhawan, P.~Żelasko, C.~Zhang, Y.-N. Chen, Y.~Tsao, J.~Balam, B.~Ginsburg, S.~M. Siniscalchi, E.~S. Chng, P.~Bell, C.~Lai, S.~Watanabe, and A.~Stolcke, ``Large language model based generative error correction: A challenge and baselines for speech recognition, speaker tagging, and emotion recognition,'' in \emph{2024 IEEE Spoken Language Technology Workshop (SLT)}, 2024, pp. 371--378.

\bibitem{busso2008iemocap}
C.~Busso, M.~Bulut, C.-C. Lee, A.~Kazemzadeh, E.~Mower, S.~Kim, J.~N. Chang, S.~Lee, and S.~S. Narayanan, ``Iemocap: Interactive emotional dyadic motion capture database,'' \emph{Language resources and evaluation}, vol.~42, pp. 335--359, 2008.

\bibitem{poria2019meldmultimodalmultipartydataset}
\BIBentryALTinterwordspacing
S.~Poria, D.~Hazarika, N.~Majumder, G.~Naik, E.~Cambria, and R.~Mihalcea, ``Meld: A multimodal multi-party dataset for emotion recognition in conversations,'' 2019. [Online]. Available: \url{https://arxiv.org/abs/1810.02508}
\BIBentrySTDinterwordspacing

\bibitem{zahiri2018emotion}
S.~M. Zahiri and J.~D. Choi, ``Emotion detection on tv show transcripts with sequence-based convolutional neural networks,'' in \emph{Workshops at the thirty-second aaai conference on artificial intelligence}, 2018.

\bibitem{10448316}
J.~Santoso, K.~Ishizuka, and T.~Hashimoto, ``Large language model-based emotional speech annotation using context and acoustic feature for speech emotion recognition,'' in \emph{ICASSP 2024 - 2024 IEEE International Conference on Acoustics, Speech and Signal Processing (ICASSP)}, 2024, pp. 11\,026--11\,030.

\bibitem{dubey2024llama3herdmodels}
\BIBentryALTinterwordspacing
A.~Dubey, A.~Jauhri, A.~Pandey, A.~Kadian, A.~Al-Dahle \emph{et~al.}, ``The llama 3 herd of models,'' 2024. [Online]. Available: \url{https://arxiv.org/abs/2407.21783}
\BIBentrySTDinterwordspacing

\bibitem{wu2022self}
Z.~Wu, Y.~Wang, J.~Ye, and L.~Kong, ``Self-adaptive in-context learning: An information compression perspective for in-context example selection and ordering,'' \emph{arXiv preprint arXiv:2212.10375}, 2022.

\bibitem{min2022rethinking}
S.~Min, X.~Lyu, A.~Holtzman, M.~Artetxe, M.~Lewis, H.~Hajishirzi, and L.~Zettlemoyer, ``Rethinking the role of demonstrations: What makes in-context learning work?'' in \emph{Proceedings of the 2022 Conference on Empirical Methods in Natural Language Processing}, 2022, pp. 11\,048--11\,064.

\bibitem{liu2021makes}
J.~Liu, D.~Shen, Y.~Zhang, B.~Dolan, L.~Carin, and W.~Chen, ``What makes good in-context examples for gpt-$3 $?'' \emph{arXiv preprint arXiv:2101.06804}, 2021.

\bibitem{reimers2019sentencebertsentenceembeddingsusing}
\BIBentryALTinterwordspacing
N.~Reimers and I.~Gurevych, ``Sentence-bert: Sentence embeddings using siamese bert-networks,'' 2019. [Online]. Available: \url{https://arxiv.org/abs/1908.10084}
\BIBentrySTDinterwordspacing

\bibitem{jiang2023mistral7b}
\BIBentryALTinterwordspacing
A.~Q. Jiang, A.~Sablayrolles, A.~Mensch, C.~Bamford, D.~S. Chaplot, D.~de~las Casas, F.~Bressand, G.~Lengyel, G.~Lample, L.~Saulnier, L.~R. Lavaud, M.-A. Lachaux, P.~Stock, T.~L. Scao, T.~Lavril, T.~Wang, T.~Lacroix, and W.~E. Sayed, ``Mistral 7b,'' 2023. [Online]. Available: \url{https://arxiv.org/abs/2310.06825}
\BIBentrySTDinterwordspacing

\bibitem{hsu2021hubert}
W.-N. Hsu, B.~Bolte, Y.-H.~H. Tsai, K.~Lakhotia, R.~Salakhutdinov, and A.~Mohamed, ``Hubert: Self-supervised speech representation learning by masked prediction of hidden units,'' \emph{IEEE/ACM transactions on audio, speech, and language processing}, vol.~29, pp. 3451--3460, 2021.

\bibitem{baevski2020wav2vec}
A.~Baevski, Y.~Zhou, A.~Mohamed, and M.~Auli, ``wav2vec 2.0: A framework for self-supervised learning of speech representations,'' \emph{Advances in neural information processing systems}, vol.~33, pp. 12\,449--12\,460, 2020.

\bibitem{radford2023robust}
A.~Radford, J.~W. Kim, T.~Xu, G.~Brockman, C.~McLeavey, and I.~Sutskever, ``Robust speech recognition via large-scale weak supervision,'' in \emph{International conference on machine learning}.\hskip 1em plus 0.5em minus 0.4em\relax PMLR, 2023, pp. 28\,492--28\,518.

\bibitem{team2024gemma}
G.~Team, M.~Riviere, S.~Pathak, P.~G. Sessa, C.~Hardin, S.~Bhupatiraju, L.~Hussenot, T.~Mesnard, B.~Shahriari, A.~Ram{\'e} \emph{et~al.}, ``Gemma 2: Improving open language models at a practical size,'' \emph{arXiv preprint arXiv:2408.00118}, 2024.

\bibitem{busso2016msp}
C.~Busso, S.~Parthasarathy, A.~Burmania, M.~AbdelWahab, N.~Sadoughi, and E.~M. Provost, ``Msp-improv: An acted corpus of dyadic interactions to study emotion perception,'' \emph{IEEE Transactions on Affective Computing}, vol.~8, no.~1, pp. 67--80, 2016.

\end{thebibliography}

\end{document}